# Symmetry-aware Depth Estimation using Deep Neural Networks

Guilin Liu[1], Chao Yang[3], Zimo Li[3], Duygu Ceylan[2], Qixing Huang[3]

[1]George Mason University, [2]Adobe Research, [3]Toyota Technological Institute at Chicago

**Abstract.** Due to the abundance of 2D product images from the internet, developing efficient and scalable algorithms to recover the missing depth information is central to many applications. Recent works have addressed the single-view depth estimation problem by utilizing convolutional neural networks. In this paper, we show that exploring symmetry information, which is ubiquitous in man made objects, can significantly boost the quality of such depth predictions. Specifically, we propose a new convolutional neural network architecture to first estimate dense symmetric correspondences in a product image and then propose an optimization which utilizes this information explicitly to significantly improve the quality of single-view depth estimations. We have evaluated our approach extensively, and experimental results show that this approach outperforms state-of-the-art depth estimation techniques.

## 1 Introduction

Internet product images (e.g., those returned from Bing and Google search engines) are becoming a popular visual medium across computer vision and graphics. They are well organized, easy to access online, exist in large quantities (counted in the billions), and provide rich and comprehensive recordings of objects in the physical world. In addition, the quality of such images is typically high, enabling us to clearly perceive the geometry and material of the underlying objects.

As 2D projections of 3D objects, however, product images lack a very important attribute: depth information. Such information is critical for many applications including image editing, image decomposition and view-independent image and shape retrieval. Unfortunately, recovering the missing depth information from product images is intrinsically hard since the underlying objects typically only appear in single images.

Depth perception is a complicated process that involves many factors [28], and is especially difficult in the single image case as it remains highly underdetermined. The heart of this difficulty lies in the fact that any number of 3-dimensional configurations can lead to the same 2-dimensional projection we see. Despite this, humans discern depth and other structural information quite easily from single images. This is only possible because of peoples' ability to infer the 3D world based on specific visual cues, such as lighting and edges. Inspired by this, and the recent success of deep learning algorithms in approaching human-level performance on a variety of tasks (e.g. image classification), we believe adopting convolutional neural networks (CNNs) for monocular depth reconstruction has vast potential. We are not the first to employ CNNs for depth reconstruction (see related works), but we found that the standard approach of utilizing CNNs as a regression tool to estimate per-pixel depth values does not yield satisfactory results.

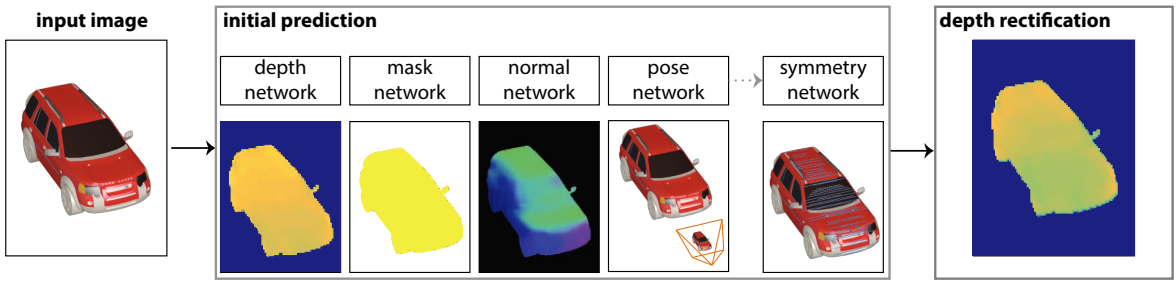

Fig. 1: We predict dense symmetry correspondences in a single product image to significantly improve the quality of depth estimation: (a) input image (b) initial depth prediction (c) predicted symmetry correspondences (only some are shown for visualization purposes) (d) refined depth estimation

To this end, we propose to explore specific image cues to improve on such results. In particular, as most product shapes admit an underlying reflectional symmetry, we use CNNs to predict dense symmetry correspondences in the given product image. We then formulate an optimization problem which utilizes this symmetry information to improve per-pixel depth predictions. We employ this optimization with per-pixel normal predictions to effectively propagate constraints from symmetric correspondences to the rest of the image.

We create a new benchmark dataset to quantitatively evaluate the performance of our approach. This benchmark consists of four classes of high-quality synthetic images. Experimental results show that learning symmetric correspondences contributes significantly towards reconstruction quality. In addition, enforcing consistency relations between the per-pixel-normal and per-pixel-depth information results in better recovery of shape details. Our method outperforms state-of-the-art methods that predict depth by using a single convolutional neural network as well as by aligning 3D shapes to the images [34].

In summary, we present the following two contributions in this paper:

- We propose a network architecture to predict dense symmetry correspondences from product images.
- We introduce an optimization that leverages the predicted symmetry correspondences and normals to improve upon state-of-the-art single-view depth estimation methods.

## 2 Related Work

The reconstruction of 3D structure from 2D data is a problem with numerous applications in recognition, segmentation and 3D modeling. While structure-from-motion and multi-view-stereo techniques are effective in this task, gleaning such information within the monocular setting is orders of magnitude more difficult. Nonetheless, show anyone a photograph, even of objects and places one has never encountered, and the 3D layout will be readily apparent - the human visual system implicitly makes heuristic assumptions about the 3D environment while also leveraging various image cues in order to accomplish this. This has inspired researchers to leverage such cues when developing algorithms for monocular depth and shape reconstruction.

**Geometric cues.** Extracted edges and contours are one of the earliest explored image cues used to infer geometric depth [29,16]. Vanishing lines constitute a special subset of these image edges that help to determine major surface orientations [21]. Representing objects as geometric primitives [3] and further exploring physical constraints among them [12] is another possibility. Assuming the 3D environment can be represented by piecewise-linear surfaces, one can interpret an image as a collection of planar patches [13,30]. A stronger assumption is that of the Manhattan-world, which conjectures that such planar patches are aligned with one of the three dominant orientations [11,39]. Even though such geometric primitives provide strong tools to tackle the ill-posed depth estimation problem, it is typically not easy to detect such primitives reliably in natural images.

**Lighting-based cues.** Shading and lighting details of an image provide a lot of geometric information. Variations in the amount of light reflected from a surface is related to the orientation of the surface relative to the light source. Earlier works that explore this tend to make strong assumptions about albedo and illumination [14], whereas more recent work employs richer priors [2]. Still, complex illumination conditions and surfaces that are not Lambertian continue to pose challenges for such methods. More recently, Shi et al. [31] estimate depth by leveraging small-scale blur effects, which exist in almost every image.

**Data-driven approaches.** Given the diversity of images in terms of content and illumination, the recent emergence of data-driven methods is not surprising. Such methods use training examples with known depth [24] or surface orientation [10,20] and transfer this knowledge to new images by utilizing pixel or patch-wise features. Recently, Choi et al.[7] proposed to transfer depth gradients instead of actual depth values to better handle the diversity between training and testing images.

Researchers have also begun to leverage 2D image and 3D shape collections for the task of depth prediction. Using keypoint correspondences between an input image of an object and a set of similar images (of not the same object), Carreira et al. [4] were able to synthesize novel virtual views and employ a structure-from-motion framework to recover the underlying 3D structure of a single-view object. Following a different vein, it is also possible to use 3D CAD models to infer depth from single images by aligning the two together [34,1]. More recently, Huang et al. [15] proposed a pipeline which uses networks of 3D shapes and 2D images simultaneously. While such data driven approaches are effective in regularizing the ill-posed depth estimation problem, they are limited by the amount of variation captured in the 2D or 3D shape collection and often fail when such collections lack objects or object parts resembling the input image.

**Deep learning methods.** The recent success of deep learning and convolutional neural networks (CNNs) has inspired researchers to consider CNNs as well for the case of monocular 3D inference. Such efforts typically first obtain a global estimate of the depth (or surface orientation) and then compute local refinements. Eigen et al. [9,8] utilize two different network architectures to obtain global and local depth estimates, while Li et al. [22] use conditional random fields (CRF) to refine the global depth estimates obtained from the neural network. Liu et al. [23] propose to directly learn the unary and potential terms of a CRF model. Wang et al. [35], on the other hand, propose a fusion network that combines the global and local surface normal estimates. These networks, however, consider the depth (or surface normal) information in isolation from all other image cues.

**Symmetry.** While our method is inspired by recent data-driven methods, we utilize an additional source of information which is ubiquitous in man-made shapes: *symmetry*. There has been a vast amount of work to detect symmetries both in 2D and 3D shapes (we refer the reader to the surveys by Liu et al. [25] and Mitra et al. [26]). In the context of utilizing symmetry information for inferring 3D structure from images, most prior work assumes symmetries are provided interactively by the user [17,37] or are detected prior to 3D structure estimation [36,19,38,33]. Shimshoni et al. [32] provide interesting results on both inferring symmetry and 3D shape starting from a manually provided symmetric pixel pair.

## 3 Problem Statement and Method Overview

In this section, we describe the problem statement and present an overview of our solution.

### 3.1 Problem Statement

Our goal is to perform depth reconstruction from a single image, $\mathcal{I}$, (with possibly cluttered background), of an object that admits a global reflection symmetry. In this paper, we fix the size of the input image to be $512 \times 512$. Without loss of generality, we assume the object is defined in a world-coordinate system, and the symmetry plane is the $yz$-plane. The output consists of (i) the object mask $\mathcal{M}$, (ii) per-pixel depth, $z_p$ for each pixel $p \in \mathcal{M}$, and (ii) symmetric pixel correspondences, $\mathcal{C} \subset \mathcal{M} \times \mathcal{M}$, established with respect to the underlying symmetry. In order to predict symmetry correspondences and propagate the detected symmetry information to the rest of the image, we also utilize a network architecture to detect the camera pose and per-pixel normal $\mathbf{n}_p$ in the world coordinate system.

We consider a 5-parameter camera model, $\mathcal{C} = (R, t_x, s)$, where $R$, $t_x$, and $s$ encode the rotation, translation, and field-of-view parameter of the camera respectively. Note that we set the translations along $y$- and $z$-axes to be zero to eliminate the redundancy in moving the reconstructed 3D object in the $yz$-plane. In our experiments, we found such a 5-parameter camera model to sufficiently represent most of the product images. We would like to note that our method can easily be extended to handle more general camera models.

Given the per-pixel depth values and the camera pose, we can easily compute the position of each pixel in the world coordinate system. Specifically, let $p \in \mathcal{I}$ be a pixel with image coordinates $(p_x^{2D}, p_y^{2D})$ and depth $z_p$ that represents the 3D point, $\mathbf{p}$, in the world coordinate system. The relation between $p$ and $\mathbf{p}$ is defined by first back-projecting the image point to the camera coordinate system (i.e. $\mathbf{p}^{cam}$) and then transforming the point in the camera coordinate system to the world coordinate system:

$$\mathbf{p} = R\mathbf{p}^{cam} + \begin{pmatrix} t_x \\ 0 \\ 0 \end{pmatrix} = z_p R \begin{pmatrix} s \cdot p_x^{2D} \\ s \cdot p_y^{2D} \\ 1 \end{pmatrix} + \begin{pmatrix} t_x \\ 0 \\ 0 \end{pmatrix}. \qquad (1)$$

### 3.2 Method Overview

Given an input product image, the proposed single-view 3D reconstruction method is divided into two stages (see Figure 1). The first initial prediction stage trains convolutional neural networks to predict the camera pose, object mask, per-pixel depth and normal, and symmetry correspondences. The second stage proposes an optimization problem that leverages the normal and symmetry correspondences to rectify the depth information.

**Initial prediction.** We adapt state-of-the art networks to obtain initial estimates of per-pixel depth information as well as object mask, camera pose and per-pixel normal. We additionally design a network architecture to predict dense symmetry correspondences. As the quality of symmetry correspondences plays a key role in our method, we first use the predicted camera pose to rectify the input image. This turns a 2D flow-field prediction into a 1D flow-field prediction problem, boosting the prediction performance.

**Depth rectification.** Once we obtain individual predictions for object mask, camera pose, per-pixel depth and normal, and symmetry correspondences, we set up an optimization problem to leverage the symmetry correspondences to rectify the predicted depth by enforcing consistency among these signals. This optimization utilizes per-pixel normal predictions to more reliably propagate the symmetry information to the whole image.

## 4  Network Design

As illustrated in Figure 1, we utilize five networks to provide initial predictions for object mask, camera pose, depth, normal, and symmetric correspondences.

We adapt state-of-the-art networks for each of these specific tasks. In the following, we highlight the modifications we made and describe the training loss for each network.

**Mask network** The mask network adapts a recent network proposed by Noh et al [27] for dense prediction of image semantic labels. It consists of standard convolution and pooling layers as well as deconvolution and unpooling layers to recover local details. We add one convolutional and deconvolution layer so that the network takes $512 \times 512$ images as input. And we use the Softmax Loss as the loss function for training.

The output of the mask network is utilized to generate a segmented image that is provided as input to the other network components and the final depth rectification optimization.

**Pose network.** The pose network adapts the architecture proposed by Kendall et al [18] for rigid pose estimation. The network replaces the three last layers of GoogleLet by a regression layer. We add one convolution layer to the input layer to fit the input image size. We also modify the regression layer so that it outputs a quaternion $\mathbf{q} \in \mathcal{R}^4$, a translation along the $x$-axis, and a field-of-view. Given a training image $I$ with ground-truth camera poses $\mathbf{q}^\star, t_x^\star, s^\star$, we define the pose loss as

$$l_{cam}(\mathcal{I}) = \|\frac{\mathbf{q}}{\|\mathbf{q}\|} - \mathbf{q}^\star\|^2 + \|t_x - t_x^\star\|^2 + (s - s^\star)^2, \qquad (2)$$

Please recall that a quaternion $\mathbf{q} = (q_r, \mathbf{q}_n)$ can be converted into a rotation matrix as:

$$R = (1 - 2\|\mathbf{q}_n\|^2)I_3 + 2\mathbf{q}_n\mathbf{q}_n^T + 2q_r(\mathbf{q}_n\times). \qquad (3)$$

**Depth and normal networks.** Eigen et al [8] proposed a network architecture that jointly infers depth and normal. We found that training the two networks separately is simpler and leads to better final results. Additionally, we ensure consistency between these signals in the depth rectification optimization performed later. For depth prediction, we use the architecture of Eigen et al [9]. For normal prediction, we use the architecture of Eigen et al [8]. For both networks we add convolutional layers to fit $512 \times 512$ images.

The loss function for depth prediction is similar to Eigen et al. [8], where we compare the log of the predicted and ground-truth per-pixel depths, $z_p, p \in I$ and $z_p^\star, p \in I$ respectively. Let $e_p = \log z_p - \log z_p^\star$, then the loss function is defined as:

$$l_{depth}(\mathcal{I}) = \frac{1}{|\mathcal{M}^\star|} \sum_{p \in \mathcal{M}^\star} e_p^2 - \frac{1}{2|\mathcal{M}^\star|^2}\Big(\sum_{p \in \mathcal{M}^\star} e_p\Big)^2. \qquad (4)$$

We use the L2-norm to penalize the deviations between the predicted normals and ground truth per-pixel normals. Let $\mathbf{n}_p^{\text{cam}}$ and $\mathbf{n}_p^{\text{cam}\star}$ be the predicted normals and the ground-truth normals in the camera coordinate system, the normal loss function is given by

$$l_{normal}(\mathcal{I}) = \frac{1}{|\mathcal{M}^\star|} \sum_{p \in \mathcal{M}^\star} \|\mathbf{n}_p^{\text{cam}} - \mathbf{n}_p^{\text{cam}\star}\|. \qquad (5)$$

Compared with the dot-product loss function proposed by Eigen et al. [8], which is less sensitive to small errors, the L2-norm leads to more accurate predictions.

**Symmetry correspondence network.** The core of the initial prediction step of our approach is the symmetry correspondence network. We adapt the FlowNet architecture, a state-of-the-art 2D flow-field prediction architecture to estimate dense symmetry correspondences in a product image. The key modification in our approach is to use the predicted camera pose to rectify the input image. This allows us to convert the 2D flow-field prediction into a 1D flow-field prediction problem, leading to significant performance gains. For image rectification, we use the method described in [6], which takes a multi-channel input image $I$, the predicted rotation $R$ and field-of-view $s$, and outputs a rectified image $I^r$ of the same size. For 1D flow-field prediction, we modify the FlowNet architecture by cutting the number of channels in each layer by half. Please refer to the supplemental material for details.

The symmetric correspondence loss is defined as

$$l_{sym}(\mathcal{I}) = \frac{1}{|\mathcal{M}_s^\star|} \sum_{p \in \mathcal{M}_s^\star} (f_p - f_p^\star)^2, \tag{6}$$

where $f_p$ is the predicted displacement for $p$ on the rectified image and $f_p^\star$ is the ground-truth displacement.

We visualized the detailed architectures of the aforementioned networks in supplemental material.

Even though this network architecture provides reasonable symmetry predictions, lack of sufficient amount of discriminative features for some input images (e.g. due to lack of texture) leads to ambiguities in predicting reliable symmetry correspondences. To filter out such wrong predictions, we perform a simple consistency check. Specifically, if the network predicts pixels $(a, b)$ and $(b, c)$ to be correspondences, we keep them only if the distance between pixel $a$ and pixel $c$ is less than a threshold (7 pixels in our experiments).

## 5   Depth Rectification

In the initial prediction step of our approach, for each input image $I$, we predict its camera parameters $R, t_x, s$, an object mask ($\mathcal{M}$), symmetric correspondences ($\mathcal{C} \subset \mathcal{M} \times \mathcal{M}$), per-pixel depth ($\overline{z}_p, p \in \mathcal{M}$) and normals ($\overline{\mathbf{n}}_p^{\text{cam}}, p \in \mathcal{M}$). Our next goal is to explore the symmetry information to improve the initial depth estimation.

Specifically, the proposed depth rectification step optimizes the depth value $z_p$ of each pixel $p \in \mathcal{M}$, so that symmetric pixels have consistent depth values. Moreover, we utilize the per-pixel normal predictions to ensure smooth depth changes across neighboring pixels. The objective function consists of three terms. The first term penalizes the deviations between the predicted and rectified depth. We employ the L1-norm to effectively handle outliers in the predicted depth values:

$$f_{\text{depth}} = \sum_{p \in \mathcal{M}} |z_p - \overline{z}_p|. \tag{7}$$

The second term enforces that for each pair of adjacent pixels $p$ and $q$, their displacement vector is perpendicular to the predicted normal at $p$. We formulate this term in the camera coordinate system and again employ the L1-norm to handle outliers:

$$f_{\text{normal}} = \sum_{p \in \mathcal{M}} \sum_{q \in \mathcal{G}_p} |(\mathbf{p}^{\text{cam}} - \mathbf{q}^{\text{cam}})^T \overline{\mathbf{n}}_p^{\text{cam}}|, \qquad (8)$$

where $\mathcal{G}$ connects for each pixel the 8 adjacent pixels.

The third term enforces the symmetric correspondences to represent consistent 3D points in the world coordinate system with respect to the global reflection plane. Denote $P = \text{diag}(-1, 1, 1)$ as the reflectional symmetry with respect to the $yz$-plane, we define the symmetry term as:

$$f_{\text{symmetry}} = \sum_{c=(p,q) \in \mathcal{S}} \|P(\mathbf{p}) - \mathbf{q}\|, \qquad (9)$$

where $P(\mathbf{p})$ denotes the reflection of $\mathbf{p}$ with respect to $P$.

Combining (7), (8) and (9) yields the following optimization problem:

$$\underset{z_p}{\text{minimize}} \sum_{p \in \mathcal{M}} f_{depth}(p) + \lambda f_{\text{normal}}(p) + \mu f_{\text{symmetry}}(p), \qquad (10)$$

where the tradeoff parameters $\lambda$ and $\mu$ are determined by applying cross-validation on %10 of the training set.

We solve (10) using re-weighted least squares, which is a popular method for optimizing objective terms under the L1-norm. We refer the reader to the supplementary material for the details of the derivations.

## 6 Experimental Results

We consider four popular object categories where the underlying reflectional symmetry is salient: chair, car, table, and sofa. An important challenge is to obtain ground-truth data to train each individual network. Standard dataset creation approaches such as human labeling or scanning are inappropriate for us due to the limitations in cost and in collecting diverse physical objects. In this paper, we propose to generate the ground-truth data through rendering ShapeNet models [5]. We employ an open-source physically-based rendering software, *Mitsuba*, to generate realistic renderings. We use $700-2500$ models for each category to generate training data. For each selected object, we choose 36 random views, each of which provides an image with ground-truth geometric information. For each training dataset, we leave out 20% of the data for validation. Figure 2 shows some example renderings.

### 6.1 Experimental Setup

We evaluate the performance of predictions for signals including camera pose, depth, and symmetry in Table 1. For camera pose estimation, we evaluate the mean errors in rotation, translation and field-of-view. Both the prediction errors in rotation and field-of-view are measured in degrees. For camera translation, we report the relative prediction error with respect to the average object size. For depth prediction, we consider the predictions at each valid pixel $p \in I$. The quality of the depth prediction is assessed by computing the relative error ($rel$) in the estimated depth $z_p$ averaged over valid pixels:

$$e_d(p) = \frac{1}{N} \sum_p |z_p - z_p^\star|/z_p^\star,$$

where $N$ denotes the total number of pixels and root mean squared error ($rms$):

$$\sqrt{\frac{1}{N} \sum_p (z_p - z_p^\star)^2}.$$

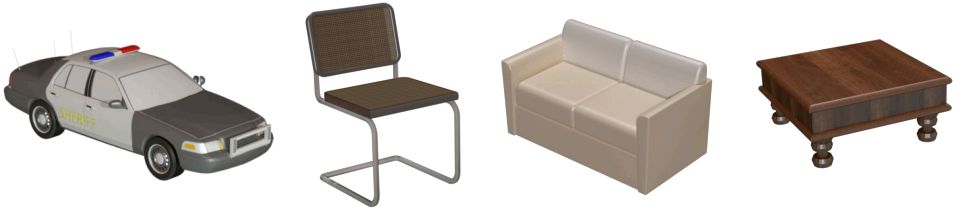

Fig. 2: A random subset of the testing images.

We also report the accuracy $\sigma$ which denotes the percentage of pixels wherein the ratio between the ground truth and predicted depth estimates is less than a threshold (1.25 and $1.25^2$).

To evaluate the quality of the symmetry prediction, we compute the average distance between the ground truth and predicted correspondences in the input image, sized at $512 \times 512$.

Figure 2 illustrates some representative results of the proposed approach. We refer to the appedix for a more complete set of results. We now summarize our findings.

**Camera prediction.** The camera network recovers the underlying camera configuration reasonably well, in particular the pose and field-of-view. This ensures that we have accurate information when rectifying images for predicting symmetric correspondences. The prediction on camera translation is less accurate. This coincides with the relative performance between translation prediction and rotation prediction using a similar architecture trained for scenes [18].

**Symmetry.** Our symmetry prediction network typically estimates the symmetry correspondences in the vicinity of the ground truth correspondence (with an average error of 5-7 pixels). We observe that reducing the search space to 1D via image rectification significantly improves the quality of the predictions.

**Depth prediction.** The result of the initial depth estimation network is only moderate. This is not surprising from the quality of the state-of-the-art results on depth estimation from scenes. Our global optimization strategy, however, significantly improves the quality by explicitly enforcing symmetry relations.

| Class | pose | | | symmetry | normal |
|---|---|---|---|---|---|
| | R | T | FOV | S | angle |
| Chair (2486) | 0.0530 | 0.641 | 0.207 | 6.0202 | 7.46 |
| Car (1484) | 0.0524 | 0.558 | 0.471 | 7.8877 | 23.34 |
| Sofa (1845) | 0.0398 | 0.518 | 0.437 | 5.1294 | 17.57 |
| Table (695) | 0.0481 | 0.385 | 0.193 | 6.6375 | 15.03 |

| Class | initial depth | | | | final depth | | | |
|---|---|---|---|---|---|---|---|---|
| | rel | rms | $\sigma < 1.25$ | $\sigma < 1.25^2$ | rel | rms | $\sigma < 1.25$ | $\sigma < 1.25^2$ |
| Chair (2486) | 0.2103 | 0.8323 | 0.5764 | 0.9234 | 0.1953 | 0.6853 | 0.6031 | 0.9627 |
| Car (1484) | 0.2186 | 0.7440 | 0.5445 | 0.9356 | 0.2156 | 0.6758 | 0.5872 | 0.9493 |
| Sofa (1845) | 0.2020 | 0.7020 | 0.5993 | 0.9719 | 0.2099 | 0.6456 | 0.6213 | 0.9735 |
| Table (695) | 0.2348 | 0.8333 | 0.5233 | 0.8860 | 0.2252 | 0.6797 | 0.6261 | 0.8907 |

Table 1: We provide the mean accuracy of predictions for each class (we provide the total number of models used for training and testing for each class in parentheses). For pose estimation, we provide the error of translation (T), rotation (R), and field of view(FOV), the last two of which are given in degrees. For symmetry detection, we provide the mean deviations in pixels. The predicted normal direction is evaluated in degrees. For initial and final depth estimation, we provide errors (rel, rms) and accuracy $\sigma < threshold$. For errors, lower numbers are better whereas for accuracy, higher numbers are better.

**Comparisons.** We have compared the proposed approach with [34]. As illustrated in Table 3, our reconstructions are quantitatively better than [34], as we are able to recover more accurate geometry both locally and globally. This is due to the fact that the performance of [34] heavily relies on retrieving shapes which are very similar to the images. This, however, is not guaranteed for most product images as there exist many more product images than shapes. In contrast, our learning based approach, which utilizes the power of generalization via convolutional neural networks, leads to significantly better results.

## 7  Conclusions and Future Work

In this paper, we have introduced a method for reconstructing a pixel-wise 3D point cloud from a single product image which admits an underlying reflectional symmetry. Our approach applies convolutional neural networks to jointly infer depth, normal and symmetric correspondences. Experimental results have shown that this joint scheme improves upon estimating individual components in isolation. In particular, learning the underlying symmetric correspondences can significantly improve depth prediction.

It is noteworthy that most shapes downloaded from the internet possess part information, as they were designed by stitching together individual model parts. In the future, it would be interesting to incorporate such information into our framework. Such structural information could be very beneficial for inferring all types of signals including depth, normal and symmetric correspondences. It also provides a way to train an end-to-end system that outputs complete CAD models by doing part-level geometry completion. Designing an end-to-end network that jointly predicts the different signals (i.e. depth, normal, and symmetry) is also an interesting research direction.

Table 2: **1D sym**: detected symmetry on rectified image; **2D sym**: transferring 1D symmetry to 2D symmetry on original image; **norm pred**: normal from normal network; **init depth**: initial depth prediction from depth network; **refined depth**: refined depth from joint optimization; **g.t. depth**: ground truth depth. The first part contains examples of synthetic images; the second part contains examples of real images.

| 1D sym | 2D sym | norm pred | init depth | refined depth | g.t. depth |
|---|---|---|---|---|---|
| 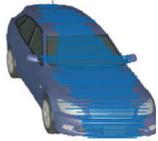 | 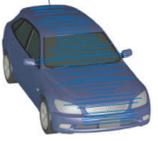 | 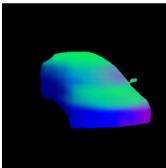 | 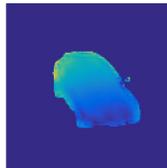 | 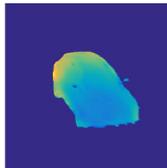 | 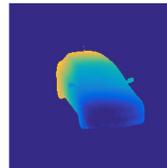 |
| 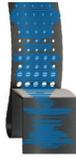 | 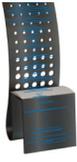 | 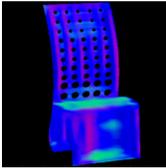 | 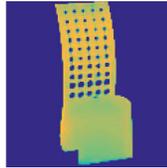 | 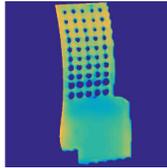 | 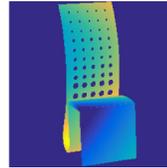 |
| 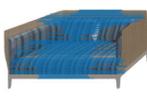 | 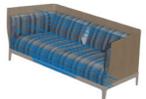 | 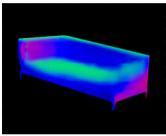 | 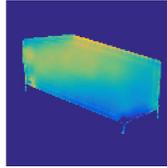 | 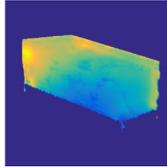 | 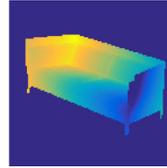 |
| 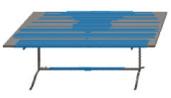 | 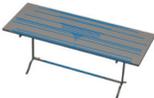 | 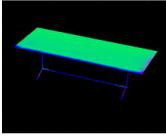 | 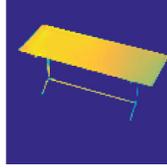 | 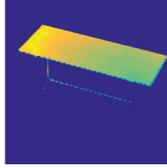 | 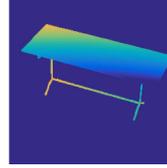 |

| 1D sym | 2D sym | norm pred | init depth | refined depth |
|---|---|---|---|---|
| 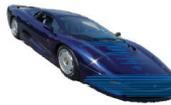 | 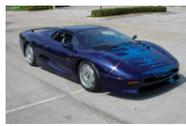 | 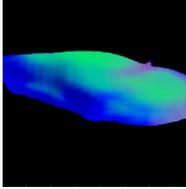 | 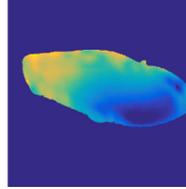 | 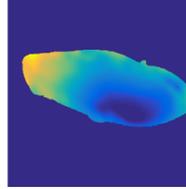 |
| 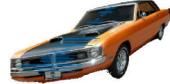 | 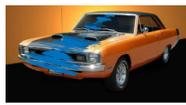 | 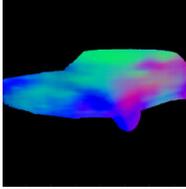 | 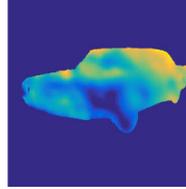 | 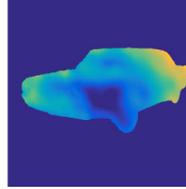 |
| 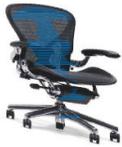 | 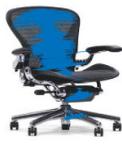 | 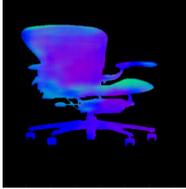 | 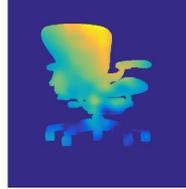 | 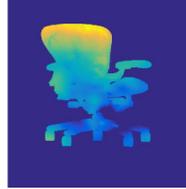 |
| 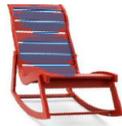 | 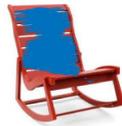 | 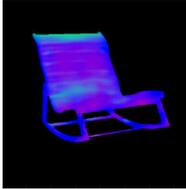 | 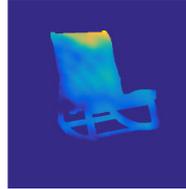 | 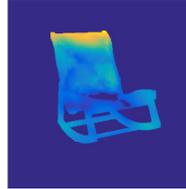 |

| Class | Ours | | | | Su et al. 14 | | | |
| --- | --- | --- | --- | --- | --- | --- | --- | --- |
| | rel | rms | $\sigma < 1.25$ | $\sigma < 1.25^2$ | rel | rms | $\sigma < 1.25$ | $\sigma < 1.25^2$ |
| Chair | 0.1953 | 0.6853 | 0.6031 | 0.9627 | 0.2924 | 0.8865 | 0.5367 | 0.9034 |
| Car | 0.2156 | 0.6758 | 0.5872 | 0.9493 | 0.2471 | 0.7921 | 0.5215 | 0.8956 |
| Sofa | 0.2099 | 0.6456 | 0.6213 | 0.9735 | 0.3011 | 0.7834 | 0.5542 | 0.7832 |
| Table | 0.2252 | 0.6797 | 0.6261 | 0.8907 | 0.2591 | 0.9341 | 0.4945 | 0.8274 |

Table 3: Comparison between our method and that of Su et al. [34]. The evaluation protocol is the same as Table 1.

## Appendix

**Depth Refinement via Gauss-Newton Optimization**

Finally, in order to discard outlier predictions, we introduce additional variable weights for each of these terms:

$$f_{\text{depth}} = \sum_{p \in \mathcal{M}} w_p |d_p - \bar{d}_p|,$$

$$f_{\text{normal}} = \sum_{p \in \mathcal{M}} \sum_{q \in \mathcal{G}_p} w_{pq} |(\mathbf{p} - \mathbf{q})^T \bar{\mathbf{n}}_p|,$$

$$f_{\text{symmetry}} = \sum_{c=(p,q) \in \mathcal{S}} w_c \|P(\mathbf{p}) - \mathbf{q}\|,$$

where each weight is initialized to be 1. We solve the weighted version of (10) in the paper iteratively. Given the weights, we first fix them and use the Gauss-Newton method to optimize for the per-pixel depths and camera parameters (see the Appendix for the details). Next, we update the weights as:

$$\begin{aligned}
w_p &= \frac{\sigma}{\sqrt{\sigma^2 + (d_p - \bar{d}_p)^2}}, \\
w_e &= \frac{\sigma}{\sqrt{\sigma^2 + \lambda((\mathbf{p} - \mathbf{q})^T \bar{\mathbf{n}}_p)^2}}, \\
w_c &= \frac{\sigma}{\sqrt{\sigma^2 + \mu \|P(\mathbf{p}) - \mathbf{q}\|^2}},
\end{aligned} \quad (11)$$

where

$$\begin{aligned}
\sigma = \text{median}(&\forall p \in \mathcal{M}, |d_p - \bar{d}_p|; \\
&\forall p \in \mathcal{M}, \forall q \in \mathcal{G}_p, \sqrt{\lambda} |(\mathbf{p} - \mathbf{q})^T \bar{\mathbf{n}}_p|; \\
&\forall (p, q) \in \mathcal{S}, \sqrt{\mu} \|P(\mathbf{p}) - \mathbf{q}\|).
\end{aligned} \quad (12)$$

This section describes the technical details of the Gauss-Newton optimization for (10) in the paper. The variables to be optimized are the intrinsic camera scaling $s$, and camera pose $R$, the camera translation along the x-axis, and the z-coordinate of each pixel $z_p, p \in \mathcal{M}$. Recall that the pixel positions and pixel depths are given by:

$$\mathbf{p} = z_p R \begin{pmatrix} sp_x^{2D} \\ sp_y^{2D} \\ 1 \end{pmatrix} + \mathbf{t}, \qquad d_p = z_p \|(1, s \cdot p_x^{2D}, s \cdot p_x^{2D})\| \quad (13)$$

The key idea of the Gauss-Newton method is to convert optimizing a non-linear least squares problem into optimizing a series of linear least squares problems. Suppose the current values of the variables are given by $z_p^c, p \in \mathcal{M}, R^c,$

$\mathbf{t}^c, s^c$, we seek to optimize a linear displacement of these variables, parameterized as follows

$$z_p = z_p^c + \delta z_p, \ R \approx (I + \mathbf{c}\times)R^c, \ \mathbf{t} = \mathbf{t}^c + \delta\mathbf{t}, \ s = s^c + \delta s. \tag{14}$$

Here $(I + \mathbf{c}\times)R^c$ is called the linear approximation of the rotation.

Substituting (14) into (13), we can write the linear approximation of $\mathbf{p}$ and $d_p$ as

$$\delta\mathbf{p} = R^c \begin{pmatrix} sp_x^{2D} \\ sp_y^{2D} \\ 1 \end{pmatrix} \delta z_p - z_p \begin{pmatrix} sp_x^{2D} \\ sp_y^{2D} \\ 1 \end{pmatrix} \times \mathbf{c}$$

$$+ z_p R^c \begin{pmatrix} p_x^{2D} \\ p_y^{2D} \\ 0 \end{pmatrix} \delta s + \delta\mathbf{t},$$

$$\delta d_p = \|(1, sp_x^{2D}, sp_y^{2D})\|\delta z_p + \frac{z_p s^c \|(p_x^{2D}, p_y^{2D})\|}{\|(1, sp_x^{2D}, sp_y^{2D})\|} \delta s. \tag{15}$$

Combining (15), (14) and (10) in the paper, we arrive at the following linear least squares formulation for optimizing the displacements:

$$\begin{aligned}
\underset{\delta z_p, \mathbf{c}, \delta\mathbf{t}, \delta s}{\text{minimize}} \quad & \sum_{p \in \mathcal{M}} w_p(d_p^c + \delta d_p - \bar{d}_p)^2 \\
& + \sum_{e=(p,q) \in \mathcal{G}} w_e\big((\mathbf{p}^c + \delta\mathbf{p} - \mathbf{q}^c - \delta\mathbf{q})^T \overline{\mathbf{n}}_p\big)^2 \\
& + \sum_{c=(p,q) \in \mathcal{S}} w_c\|P(\mathbf{p} + \delta\mathbf{p}) - (\mathbf{q} + \delta\mathbf{q})\|^2.
\end{aligned} \tag{16}$$

It is clear that the optimal values of $\delta z_p, \delta s, \mathbf{c}$ and $\delta\mathbf{t}$ can be obtained by solving a linear system. The variables at the next iteration are then given by:

$$z_p \leftarrow z_p + \delta z_p, \quad s \leftarrow s + \delta s, \quad R = \exp(\mathbf{c}\times)R^c, \quad \mathbf{t} \leftarrow \mathbf{t} + \delta\mathbf{t}. \tag{17}$$

where

$$\exp(\mathbf{c}\times) = I + \frac{\sin(\|\mathbf{c}\|)}{\|\mathbf{c}\|}(\mathbf{c}\times) + \frac{1 - \cos\|\mathbf{c}\|}{\|\mathbf{c}\|^2}(\mathbf{c}\times)^2.$$

is the exponential map.

**Rectification and De-rectification**

Suppose the rotation is given by

$$R = \begin{pmatrix} \mathbf{r}_1 \\ \mathbf{r}_2 \\ \mathbf{r}_3 \end{pmatrix}.$$

The homogeneous coordinates of the two vanishing points are given by

$$\mathbf{v}_1 = \frac{1}{s}\frac{1}{(\mathbf{r}_2 \times \mathbf{r}_3)_z}(\mathbf{r}_2 \times \mathbf{r}_3)$$
$$\mathbf{v}_2 = \frac{1}{s}\frac{1}{(\mathbf{r}_1 \times \mathbf{r}_3)_z}(\mathbf{r}_1 \times \mathbf{r}_3). \qquad (18)$$

The homography transformation is given by

$$H = \begin{pmatrix} 1 & 0 & 0 \\ 0 & 1 & 0 \\ l_a & l_b & l_c \end{pmatrix}.$$

where $(l_a, l_b, l_c)^T = \mathbf{v}_1 \times \mathbf{v}_2$.

**Synthetic Example**

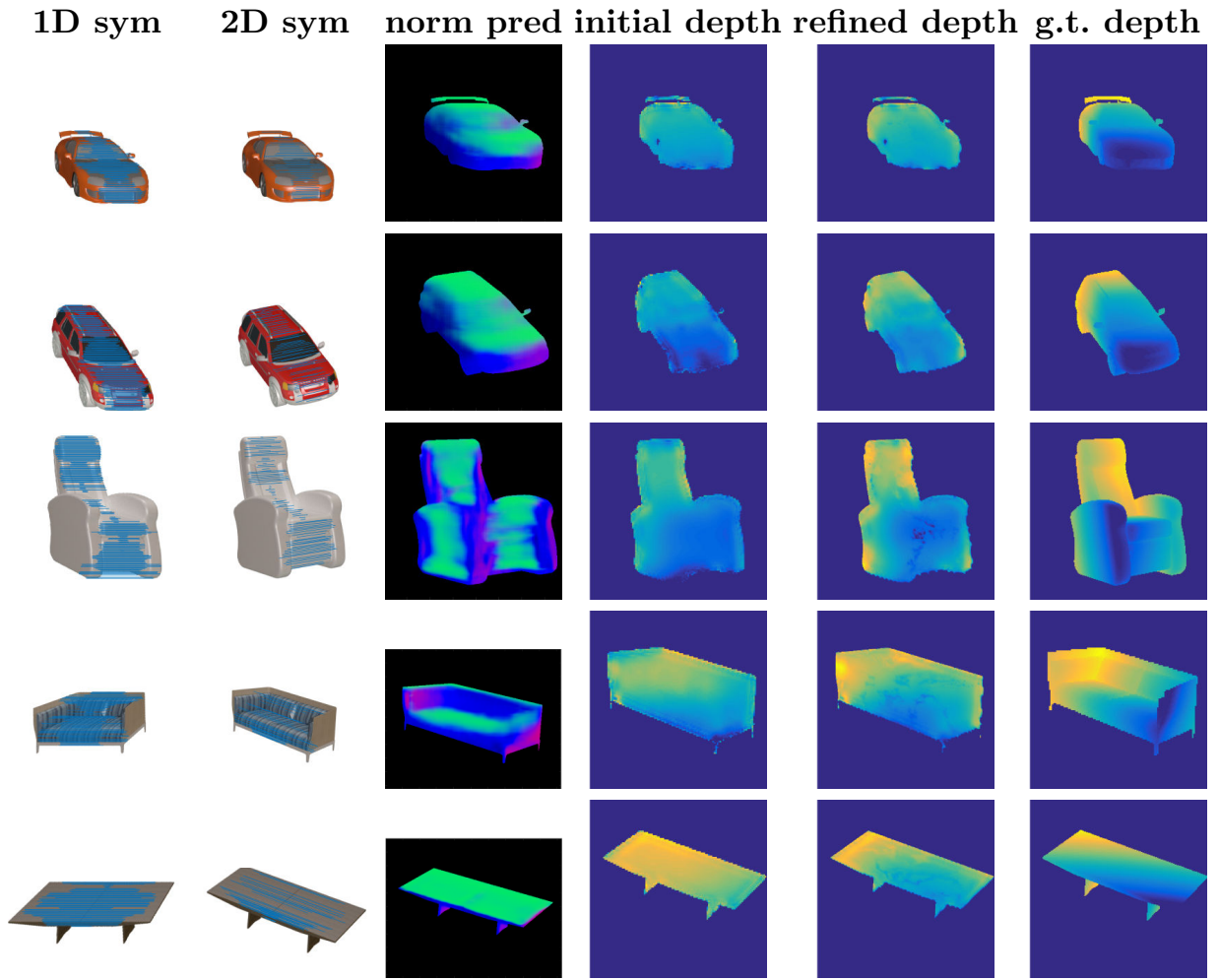

**Real Image and Product Image Example**

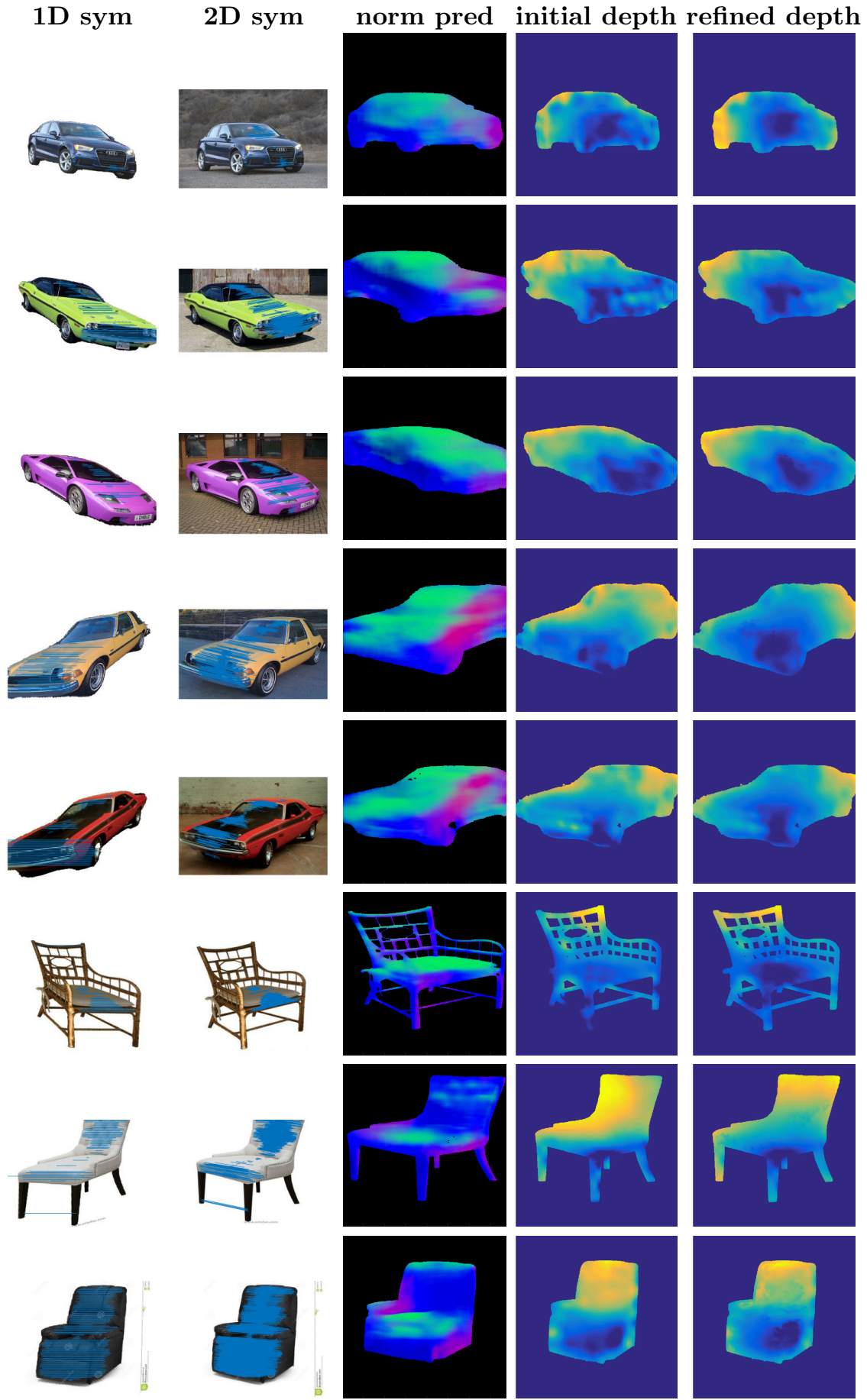

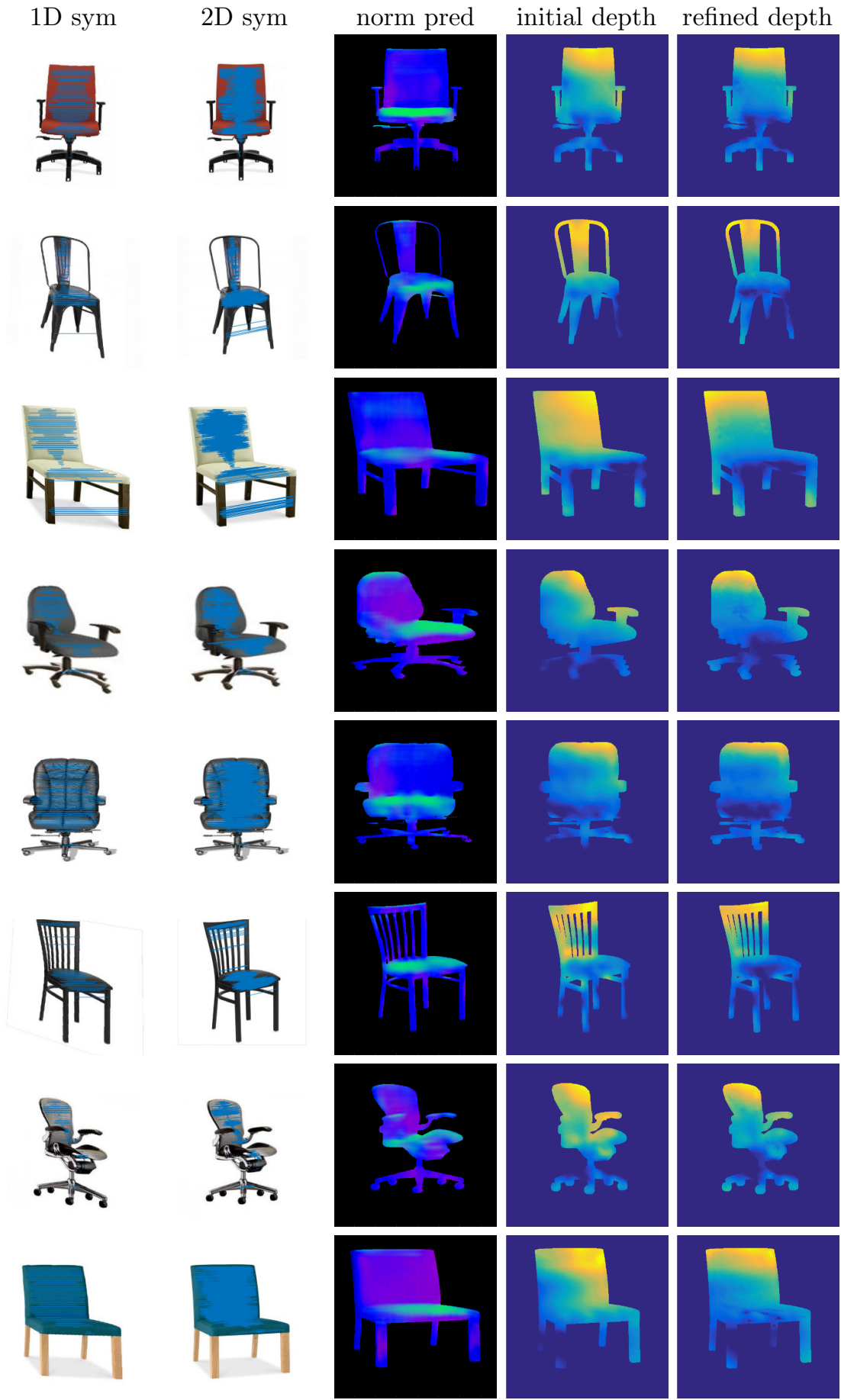